\documentclass[a4paper,fleqn]{cas-dc}
\usepackage[numbers,sort&compress,super]{natbib}
\usepackage{placeins}
\usepackage{textcomp}
\usepackage{tikz}
\usepackage{forest}
\usepackage{xcolor}
\usepackage{caption}
\usepackage{float}
\usepackage{amsmath,amssymb,amsfonts}
\usepackage{nccmath}
\usepackage{multirow}
\usepackage{dsfont} 
\usepackage{placeins}
\usepackage{booktabs,tabularx,makecell,array}
\usepackage{amssymb} % for \checkmark
\usepackage{stfloats}
\usepackage{pifont}
\restylefloat{figure}
\def\tsc#1{\csdef{#1}{\textsc{\lowercase{#1}}\xspace}}
\tsc{WGM}
\tsc{QE}
\tsc{EP}
\tsc{PMS}
\tsc{BEC}
\tsc{DE}
\begin{document}
\let\WriteBookmarks\relax
\def\floatpagepagefraction{1}
\def\textpagefraction{.001}
\shorttitle{A Discrete Diffusion Framework for Radiology Report Generation}
\shortauthors{S. Zhou et~al.}

\title [mode = title]{DRRG: A Discrete Diffusion Framework for Radiology Report Generation}                      
% \tnotemark[1,2]

% \tnotetext[1]{This document is the results of the research
%    project funded by the National Science Foundation.}

% \tnotetext[2]{The second title footnote which is a longer text matter
%    to fill through the whole text width and overflow into
%    another line in the footnotes area of the first page.}

\author[1]{Shaoyang Zhou}
% \cormark[1]
\fnmark[1]
\ead{szho0189@uni.sydney.edu.au}
% \ead[url]{www.jkkrishnan.in}

\credit{Conceptualization, Methodology, Software, Data curation, Formal analysis, Validation, Visualization, Writing -- review \& editing, Writing -- original draft}

%\address[1]{, Street 129, 1043 NX Amsterdam, The Netherlands}
\affiliation[1]{organization={School of Electrical and Computer Engineering, The University of Sydney},
                city={Sydney},
                state={NSW},
                postcode={2006},
                country={Australia}}

\affiliation[2]{organization={Department of Radiology, The First Affiliated Hospital of Soochow University},
                city={Suzhou},
                state={Jiangsu},
                postcode={215006},
                country={China}}

\affiliation[3]{organization={School of Computer Science, The University of Adelaide},
                city={Adelaide},
                state={SA},
                postcode={5005},
                country={Australia}}

\affiliation[4]{organization={School of Computing and Information Technology, University of Wollongong},
                city={Wollongong},
                state={NSW},
                postcode={2522},
                country={Australia}}
            
\author[1]{Yingshu Li}
% \cormark[1]
\fnmark[1]
\ead{yingshu.li@sydney.edu.au}
\credit{Conceptualization, Methodology, Software, Data curation, Validation, Writing -- review \& editing}

\author[1]{Yunyi Liu}
% \cormark[1]
\fnmark[1]
\ead{yunyi.liu1@sydney.edu.au}
\credit{Data curation, Validation, Investigation, Writing -- review \& editing}

\author[2]{Lijun Pu}
% \cormark[1]
\ead{20255232248@stu.suda.edu.cn}
\credit{Investigation, Validation, Methodology, Writing -- review \& editing}

\author[3]{Lingqiao Liu}
% \cormark[1]
\ead{lingqiao.liu@adelaide.edu.au}
\credit{Methodology, Investigation, Data curation, Writing -- review \& editing}

\author[4]{Lei Wang}
% \cormark[1]
\ead{leiw@uow.edu.au}
\credit{Conceptualization, Methodology, Validation, Writing -- review \& editing}

\author[1]{Luping Zhou}[orcid=0000-0001-8762-2424]
\cormark[1]
\ead{luping.zhou@sydney.edu.au}

\credit{Conceptualization, Methodology, Investigation, Data curation, Writing -- review \& editing}

\cortext[cor1]{Corresponding author.}

% \cortext[cor2]{Principal corresponding author}
\fntext[fn1]{Co-first authors.}
\begin{abstract}
\noindent\textbf{Purpose:}
Automatic radiology report generation (RRG) has been widely explored to improve reporting accuracy and reduce radiologists' workload. Most existing methods rely on autoregressive (AR) frameworks that generate reports token by token and cannot revise earlier content, making them prone to error propagation and inconsistent with the iterative refinement process of radiological reporting. In contrast, discrete diffusion large language models (DLLMs) generate text through iterative denoising, naturally enabling report refinement. However, DLLMs have not been extensively investigated for RRG. In this study, we developed and evaluated a discrete diffusion framework for RRG that enables iterative refinement rather than conventional left-to-right autoregressive decoding.

\noindent\textbf{Materials and methods:}
We developed DRRG, a DLLM-based framework that formulates RRG as iterative masked-token denoising. DRRG incorporates a clinical-entities-aware complementary mask to improve token supervision coverage and emphasize clinically important entities, together with a concept-conditioning module that injects image-derived clinical concepts into visual representations. DRRG was trained and evaluated on MIMIC-CXR and CheXpert Plus.

\noindent\textbf{Results:}
On MIMIC-CXR, DRRG achieved BLEU-4 of 0.210, CheXpert-F1 of 0.549, RadGraph-F1 of 0.281, GREEN of 0.360, and RaTEScore of 0.604, outperforming the compared methods on most reported metrics, despite employing a substantially smaller LLM decoder. On CheXpert Plus, DRRG achieved the highest BLEU-4 (0.119) and CheXpert-F1 (0.347) among the compared methods.

\noindent\textbf{Conclusion:}
Discrete diffusion provides an effective alternative to autoregressive radiology report generation by enabling iterative, bidirectional report refinement. Incorporating clinically focused masking and image-derived concept conditioning improves report quality and clinical consistency.
\end{abstract}

% \centering
% \includegraphics[
%     width=\textwidth,
%     height=0.75\textheight,
%     keepaspectratio
% ]{Graph_Abstract.png}

% \begin{highlights}
% \item DRRG generates radiology reports through iterative discrete denoising.

% \item A clinical-entities-aware complementary mask emphasizes core findings.

% \item Image-derived clinical concepts guide report denoising.

% \item Extensive evaluations validate DRRG across diverse metrics.
% \end{highlights}


\maketitle

\section{Introduction}
\subsection{Background}
Radiology reports summarize imaging findings and provide critical evidence for clinical diagnosis and treatment planning. Unlike general descriptive text, radiology reporting is safety-critical: incorrect negation or contradictory findings may lead to misinterpretation and inappropriate patient management. Therefore, semantic consistency and logical correctness are essential not only for fluency, but also for clinical reliability. Nevertheless, producing such high-quality reports is time-consuming and places a substantial burden on radiologists~\cite{cao2023current,bailey2022understanding}, especially with the rapid growth of imaging volume. Consequently, automatic radiology report generation (RRG) has received increasing attention as a means to reduce radiologists' workload~\cite{najdenkoska2022uncertainty}.

Existing RRG methods are predominantly built upon autoregressive (AR) decoding for its strong language modeling capability and have achieved remarkable progress~\cite{li2024kargen,wang2023r2gengpt}. However, this paradigm introduces potential risks for safety-critical clinical reporting, because reports are generated and fixed sequentially from left to right, with each token conditioned only on previously generated tokens. As a result, once an erroneous token is produced, it becomes part of the context for subsequent predictions, which may amplify early mistakes~\cite{duany}. This mechanism is also misaligned with real-world radiology workflow, where reports are typically drafted, reviewed, and iteratively revised to ensure coherence and correctness~\cite{sharpe2012radiology}.

\begin{figure}[!t]
\centering
   \includegraphics[width=\linewidth]{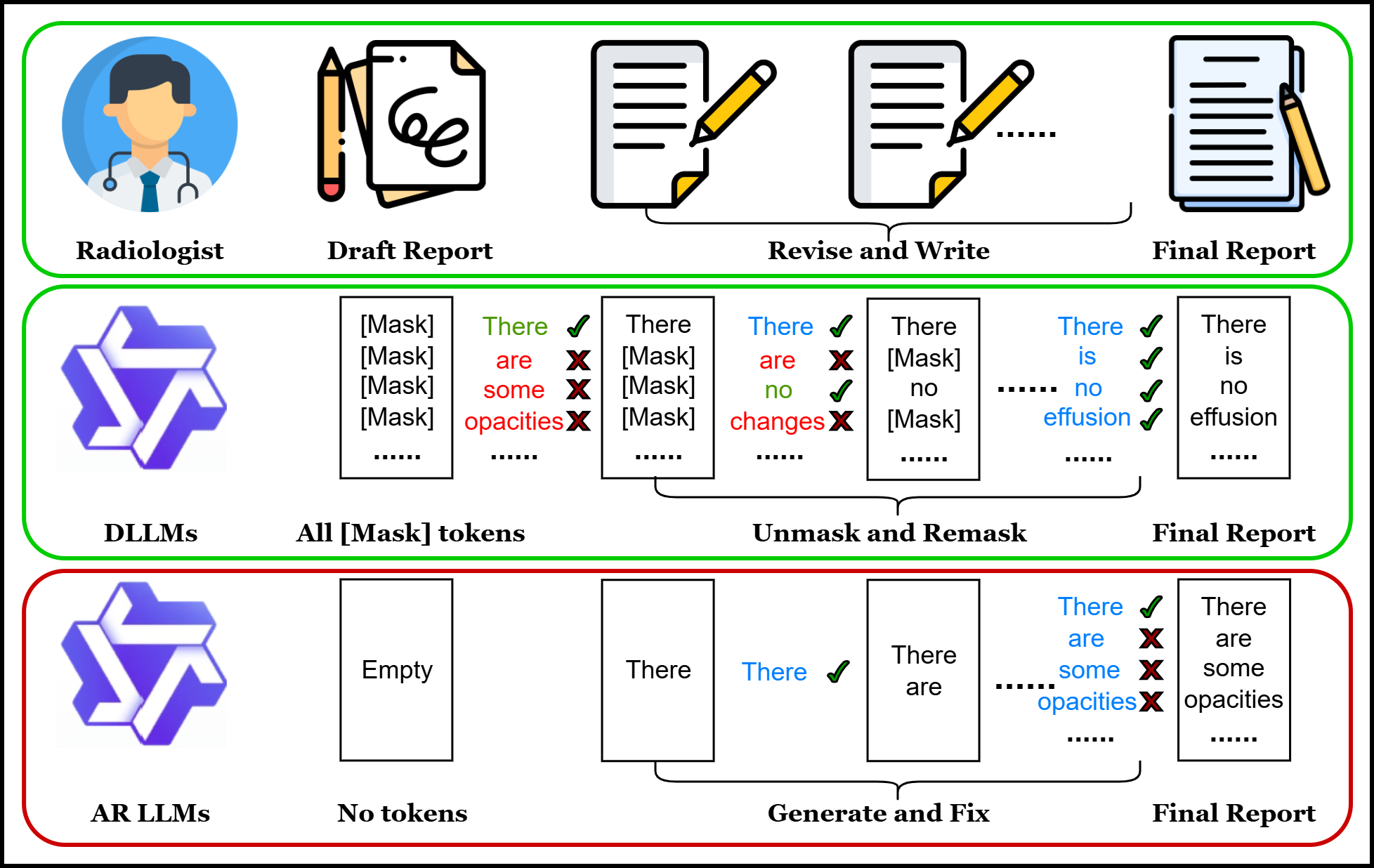}
   \caption{Comparison of radiologists' report writing and the generation patterns of DLLMs and AR LLMs. Green, blue, and red tokens denote high-confidence, fixed, and low-confidence predictions, while \checkmark{} and $\times$ indicate correct and incorrect predictions.}
\label{fig:introduction}
\end{figure}

These observations raise a fundamental question: Is strictly left-to-right generation suitable for clinical reporting? We hypothesize that radiology report generation is not merely sequential language modeling, but a global consistency optimization problem requiring report-level coordination of clinical findings. Since left-to-right decoding fixes early predictions without later revision, iterative refinement with full bidirectional context provides a more suitable paradigm for satisfying clinical constraints. Discrete diffusion large language models (DLLMs) naturally align with this process, as illustrated in Figure~\ref{fig:introduction}. Through iterative masked-token denoising, DLLMs~\cite{nie2026large} enable non-causal, bidirectional refinement conditioned on global context. Recent studies show that DLLMs can match strong AR baselines on general-domain language tasks~\cite{you2025llada,ye2025dream}. However, diffusion-based generation for RRG remains relatively underexplored, particularly regarding its ability to preserve report-level semantic consistency and accurately model specialized radiological terminology.

To investigate the applicability of discrete diffusion models to radiology report generation, we introduce DRRG, a task-specific discrete diffusion large language model. DRRG formulates report generation as iterative masked-token denoising, enabling non-causal refinement over the full report context and better aligning with the revisable nature of radiology reporting. To improve clinical reliability, we introduce a clinical-entities-aware complementary mask that enhances training effectiveness and emphasizes fine-grained clinical findings. We further propose a concept-conditioning module that injects clinically meaningful signals into denoising, promoting clinical grounding and observation-level consistency.

The key contributions of our work are summarized in the following points:
\begin{itemize}
\item We develop DRRG, a discrete diffusion framework for radiology report generation that generates reports through iterative masked-token denoising, enabling predictions to be progressively refined using bidirectional report context.

\item We introduce a clinical-entities-aware complementary mask and a concept-conditioning module. The former improves training effectiveness by increasing supervision coverage and placing greater emphasis on clinically important entities, while the latter incorporates image-derived clinical information to strengthen guidance during denoising.

\item Extensive experiments on different datasets show that DLLM-based RRG achieves performance comparable to strong AR baselines, while ablation studies and case analyses validate the proposed components and the feasibility of DLLM-based RRG.

\end{itemize}

\subsection{Related Work}
\subsubsection{Automated Radiology Report Generation}
Automated radiology report generation (RRG) aims to translate medical images into diagnostically meaningful reports and is commonly formulated as a clinically grounded extension of image captioning~\cite{vinyals2015show, you2016image}. Most existing methods adopt an encoder--decoder framework, where visual representations are extracted by an image encoder and then decoded into reports. Early CNN--RNN models~\cite{jing2019show, jing2018automatic, zhang2020radiology} were limited in modeling global context and long-range dependencies, while transformer-based methods~\cite{chen2020generating,you2021aligntransformer} improved both visual and textual representation learning through attention mechanisms~\cite{vaswani2017attention}. More recently, LLM-based decoders~\cite{wang2023r2gengpt,liu2024bootstrapping} have been introduced to transfer linguistic knowledge and domain priors to RRG, improving report fluency and factual consistency. However, despite these advances, most methods~\cite{li2025s} still rely on autoregressive decoding, which may suffer from error accumulation and limited revisability. Alternative generation paradigms that more closely reflect the iterative nature of radiology reporting have received relatively limited attention.

\begin{figure*}[!t]
   \includegraphics[width=\textwidth]{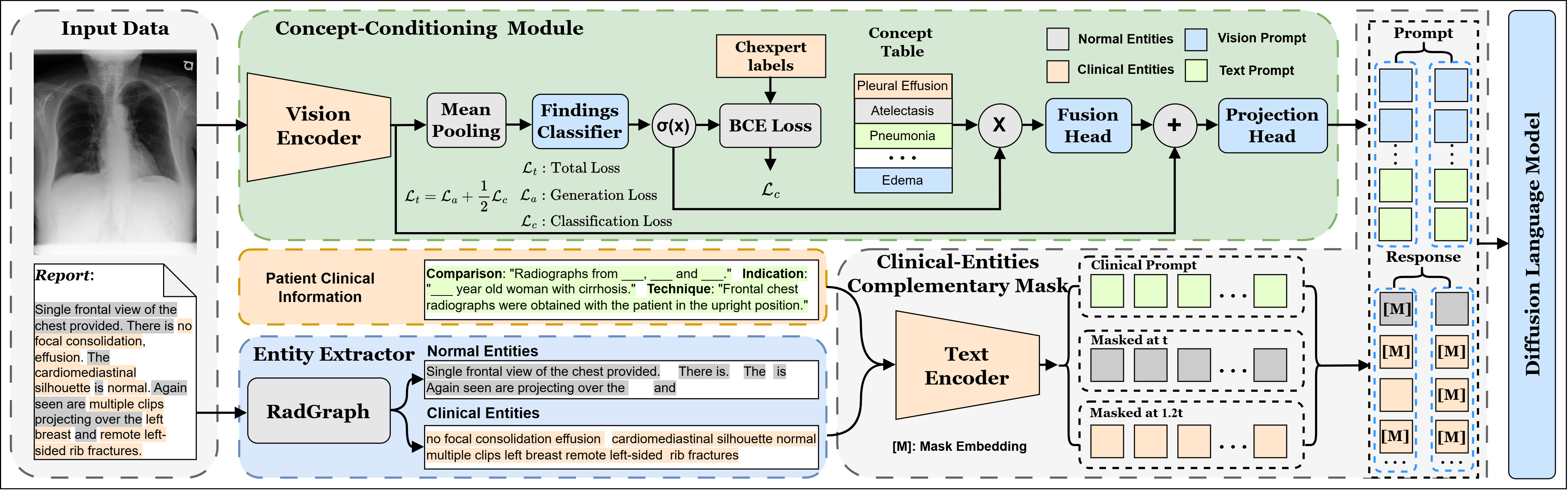}
   \caption{Overview of DRRG. Given a CXR image--report pair, DRRG extracts concept-conditioned visual features and applies clinical-entities-aware complementary masking to emphasize clinical entities. The discrete diffusion language model reconstructs masked report tokens from prompt and noise-injected response tokens, while prompt and unmasked response tokens are excluded from the loss.
}
\label{fig:method}
\end{figure*}

\subsubsection{Discrete Diffusion Language Model}
Recent DLLMs~\cite{nie2026large,ye2025dream} have shown competitive performance against AR models.
Beyond unimodal settings, subsequent efforts have extended DLLMs to multimodal architectures~\cite{you2025llada,li2025lavida}, where visual inputs are incorporated as additional conditioning signals to guide denoising and generation. Recently, LLaDA-MedV~\cite{dong2025llada} further adapts this paradigm to biomedical imagery, indicating that DLLMs can be effectively generalized to medically grounded vision--language scenarios. Despite these developments, DLLM frameworks remain relatively underexplored in the context of RRG, where generation requires clinically faithful long-form narratives and fine-grained alignment with subtle evidence.

\section{Materials and methods}
We introduce DRRG, a multimodal discrete diffusion framework for CXR report generation, as illustrated in Figure~\ref{fig:method}. This section describes the data collection and methodology of DRRG.
\label{sec:method}

\subsection{Data collection}
We evaluate our model with two datasets: MIMIC-CXR~\cite{johnson2019mimic} and CheXpert Plus~\cite{chambon2024chexpert} for radiology report generation.
\subsubsection{MIMIC-CXR} MIMIC-CXR~\cite{johnson2019mimic} is a public chest radiography dataset comprising 377,110 CXR images and 227,835 free-text reports from 64,588 patients. Following the split of~\cite{chen2020generating}, we use 270,790 training samples, 2,130 validation samples, and 3,858 test samples.

\subsubsection{CheXpert Plus}CheXpert Plus~\cite{chambon2024chexpert} is a radiology benchmark comprising 223,228 CXR images and 187,711 reports from 64,725 patients. Following the CXPMRG-Bench split~\cite{wang2025cxpmrg}, we use 40,463 training samples, 5,780 validation samples, and 11,562 test samples.

\subsection{Methodology}
\subsubsection{Model Architecture and Loss Function}
DRRG extracts visual features, derives clinical concept representations, and generates reports through a discrete diffusion language model trained with a clinical-entities-aware complementary mask. We first describe the model architecture and training objective, followed by the training strategy and inference procedure.

\textbf{Vision Encoder:}
We adopt SigLIP2~\cite{tschannen2025siglip} as the vision encoder $E$ for its strong visual representation capacity and large-scale image--text pretraining. Given a CXR image $I$, the encoder extracts a sequence of visual tokens:
\begin{equation}
V = E(I; \theta_v), \quad V \in \mathbb{R}^{B \times y \times D},
\end{equation}
where $B$ denotes the batch size, $y$ the number of visual tokens, $D$ the hidden dimension of the vision encoder, and $\theta_v$ the parameters of $E$. The visual tokens $V$ are then passed to the concept-conditioning module.

\textbf{Concept-Conditioning Module:}
Accurate noise prediction requires informative conditioning. However, visual features alone may not provide sufficiently precise guidance for the denoising process. We therefore introduce a concept-conditioning module that augments visual representations with clinical priors.

Given visual features $V \in \mathbb{R}^{B \times y \times D}$, we first apply mean pooling over the token dimension to obtain a global image representation $V_{\text{mean}} \in \mathbb{R}^{B \times D}$. A linear classifier $W_c$ then predicts $C=14$ clinical observations~\cite{irvin2019chexpert}, followed by a sigmoid activation to produce multi-label concept scores:
\begin{equation}
p_0 = \sigma(W_c(V_{\text{mean}})), \quad p_0 \in \mathbb{R}^{B \times C}.
\label{eq:concept_scores}
\end{equation}

To transform the predicted distribution into a concept representation, we introduce a learnable concept embedding table $T \in \mathbb{R}^{C \times D}$, where each row corresponds to a clinical concept. The predicted scores are used to aggregate concept embeddings through a probability-weighted combination:
\begin{equation}
e_{con} = \sum_{c=1}^{C} p_{0,c} T_c, \quad e_{con}\in \mathbb{R}^{B \times D}.
\label{eq:mul}
\end{equation}

The concept feature $e_{con}$ is further transformed by a fusion head $W_f$, broadcast along the visual-token dimension, and added to the original visual features:
\begin{equation}
\tilde{H} = V + \operatorname{Broadcast}(W_f(e_{con})),
\quad \tilde{H} \in \mathbb{R}^{B \times y \times D}.
\label{eq:concept_fusion}
\end{equation}

Finally, a MLP mapper projects $\tilde{H}$ into the text hidden-state space of dimension $d$, yielding the final conditioning representations $H \in \mathbb{R}^{B \times y \times d}$. By injecting structured clinical priors into visual tokens, this module provides clinically grounded conditioning for the denoising process.

\textbf{Discrete Diffusion Language Model:}
We adopt Qwen3-0.6B-diffusion-mdlm-v0.1~\cite{zhou2026dllm} as the backbone of our language model. It is initialized from the pretrained Qwen3-0.6B foundation model~\cite{yang2025qwen3} and further adapted with a discrete diffusion modeling objective. Its 0.6B parameter scale offers a practical balance between modeling capacity and computational efficiency.

The input sequence is constructed by concatenating the concept-conditioned visual features $H$, clinical context $Y$, including Indication, Comparison, Technique, and History, and the ground-truth report $G_0$.

Following the LLaDA training paradigm~\cite{nie2026large}, we sample a mask ratio $t \sim \mathcal{U}(0,1)$ for each training instance. Tokens in the target report region $G_0$ are independently replaced with a special \texttt{[MASK]} token with probability $t$ and kept unchanged with probability $1-t$, producing a masked sequence $G_t$. This defines the forward noising process.

\textbf{Loss Function:}
Given the masked report $G_t$, the model is trained to recover the clean report $G_0$ given the concept-conditioned features $H$ and clinical context $Y$. Following discrete diffusion modeling objectives~\cite{nie2026large}, we optimize:

\begin{equation}
\mathcal{L}_{a}
=
-\mathbb{E}_{t,H,Y,G_0,G_t}
\left[
\frac{1}{t}
\sum_{j\in\mathcal{M}_t}
\log p_{\theta}
\left(
G_0[j] \mid H,Y,G_t
\right)
\right],
\end{equation}
where $\mathcal{M}_t=\{j \mid G_t[j]=\texttt{[MASK]}\}$ denotes the set of masked positions within the response token sequence.

We further supervise the concept-conditioning module with clinical labels $y_c$ extracted by CheXbert~\cite{smit2020chexbert}. The predicted observation probabilities $p_0$ are optimized with a binary cross-entropy loss:
\begin{equation}
\mathcal{L}_{c}
= - \frac{1}{C} \sum_{c=1}^{C} 
\left[ y_c \log p_{0,c} + (1 - y_c) \log (1 - p_{0,c}) \right].
\end{equation}
The overall training objective is defined as:
\begin{equation}
\mathcal{L}_{t}=\mathcal{L}_{a}+\lambda \mathcal{L}_{c},
\end{equation}
where $\lambda$ controls the contribution of concept supervision and is set to $0.5$ in all experiments.

\begin{figure}[!t]
\centering
   \includegraphics[width=\linewidth]{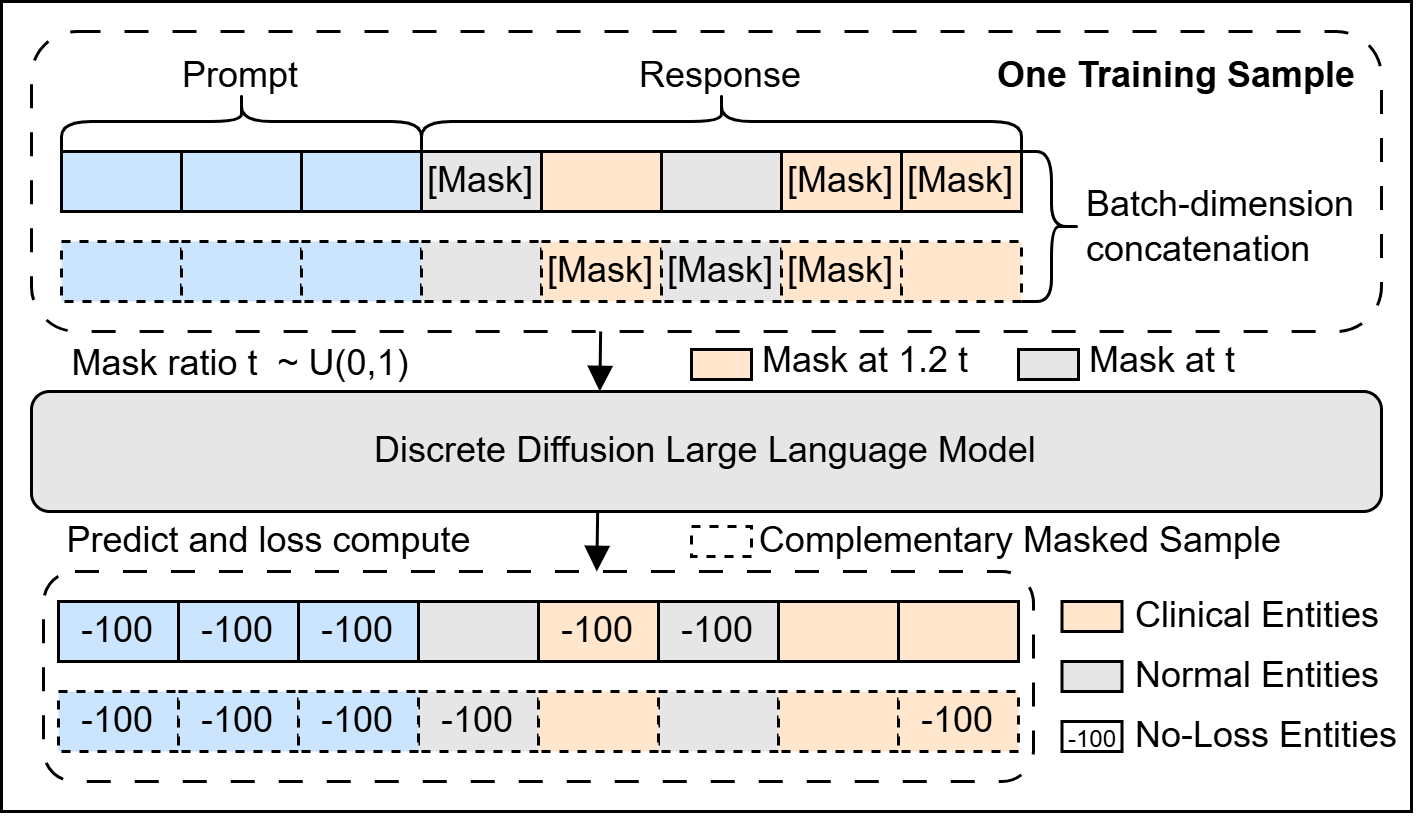}
   \caption{Illustration of the clinical-entities-aware complementary mask. Clinical entities are assigned higher masking probabilities than normal entities in both the original and complementary masked samples. The loss is computed only on masked response tokens.
}
\label{fig:mask}
\end{figure}

\subsubsection{Training Strategies and Algorithms}

\textbf{Two-stage Training:}
We apply two-stage training for cross-modal alignment. In the first stage, we freeze the visual encoder and language model, and optimize only the mapper to align visual representations with the text embedding space, thereby stabilizing cross-modal alignment. In the second stage, all modules are unfrozen and jointly fine-tuned end-to-end, allowing the entire model to adapt to the target generation task.

\textbf{Clinical-Entities-Aware Complementary Mask:}
Uniform random masking in discrete diffusion modeling is inefficient, as each forward pass supervises only a subset of tokens~\cite{li2025lavida} and treats clinically important and auxiliary tokens equally. This is particularly limiting for radiology reports, where clinical entities carry dense diagnostic information and determine report correctness.

We therefore extend complementary masking~\cite{li2025lavida} with clinical-entity awareness, as illustrated in Figure~\ref{fig:mask}. For each report $G_0$, we construct two complementary masked views, such that tokens preserved in one view are preferentially masked in the other to increase token-level supervision coverage, while assigning higher masking probabilities to clinical entity tokens to emphasize clinically important content. For normal entity tokens, the two views are masked with probabilities $t$ and $1-t$, respectively. For clinical entity tokens, we use RadGraph~\cite{jain2021radgraph} to identify entity spans in the report and assign them higher masking probabilities, i.e., $\alpha t$ and $\alpha(1-t)$ for the two views, where $\alpha$ is set to $1.2$. The entity-level mask probabilities are clipped to the range $[0,1]$ when they exceed 1. This design encourages the model to focus denoising supervision on core clinical findings while preserving complete report-level coverage.

\begin{table*}[!htbp]
\centering
\caption{NLG performance of DRRG and existing RRG methods on the MIMIC-CXR dataset. For each metric, the best-performing result is shown in bold, and the second-best result is underlined. Model names marked with \textsuperscript{*} indicate results reported in their original papers. For models using LLM-based text decoders, the corresponding parameter sizes are reported.}
\label{tab:mimic_nlg}
\scriptsize
\setlength{\tabcolsep}{2pt}
\renewcommand{\arraystretch}{0.5}
\resizebox{\textwidth}{!}{%
\begin{tabular}{@{}lllccccc@{}}
\toprule
\textbf{Dataset} & \textbf{Model} & \textbf{Publication}&
\textbf{BLEU-1} & \textbf{BLEU-2} & \textbf{BLEU-3} &
\textbf{BLEU-4} & \textbf{ROUGE-L} \\
\midrule

\multirow{28}{*}{\raisebox{-1.8ex}{\textbf{MIMIC-CXR}}}
& R2Gen\textsuperscript{*}~\cite{chen2020generating} & EMNLP'20
& 0.353 & 0.218 & 0.145 & 0.103 & - \\

& R2GenCMN\textsuperscript{*}~\cite{chen2022cross} & ACL-IJCNLP'21
& 0.353 & 0.218 & 0.148 & 0.106 & - \\

& METransformer\textsuperscript{*}~\cite{wang2023metransformer} & CVPR'23
& 0.386 & 0.250 & 0.169 & 0.124 & 0.291 \\

& DCL\textsuperscript{*}~\cite{li2023dynamic} & CVPR'23
& - & - & - & 0.109 & 0.284 \\

& KiUT\textsuperscript{*}~\cite{huang2023kiut} & CVPR'23
& 0.393 & 0.243 & 0.159 & 0.113 & 0.285 \\

& CvT2DistilGPT2\textsuperscript{*}~\cite{nicolson2023improving} & AIM'23
& 0.393 & 0.248 & 0.171 & 0.127 & 0.155 \\

& R2GenGPT\textsuperscript{*} (7B)~\cite{wang2023r2gengpt} & Meta-Rad'23
& 0.411 & 0.267 & 0.186 & 0.134 & 0.297 \\

& Bootstrapping\textsuperscript{*} (14.2B)~\cite{liu2024bootstrapping} & AAAI'24
& 0.402 & 0.262 & 0.180 & 0.128 & 0.291 \\

& EKAGen\textsuperscript{*}~\cite{bu2024instance} & CVPR'24
& 0.419 & 0.258 & 0.170 & 0.119 & 0.287 \\

& Multi-Grained\textsuperscript{*}~\cite{liu2024multi} & TMI'24
& 0.406 & 0.267 & 0.190 & 0.141 & 0.309 \\

& DAMPER\textsuperscript{*}~\cite{huang2025damper} & AAAI'25
& 0.402 & 0.284 & 0.227 & \underline{0.193} & 0.301 \\

& DART\textsuperscript{*}~\cite{Park_2025_CVPR} & CVPR'25
& 0.437 & 0.279 & 0.191 & 0.137 & 0.310 \\

& MultiP-R2Gen\textsuperscript{*} (7B)~\cite{chen2025enhancing} & TMI'25
& 0.425 & 0.279 & 0.194 & 0.140 & 0.307 \\

& KACL\textsuperscript{*} (8B)~\cite{sha2025contrastive} & MICCAI'25
& 0.414 & 0.270 & 0.184 & 0.136 & 0.303 \\

& MambaXray-VL-Large\textsuperscript{*} (7B)~\cite{wang2025cxpmrg} & CVPR'25
& 0.422 & 0.268 & 0.184 & 0.133 & 0.289 \\

& REVTAF-RRG\textsuperscript{*}~\cite{zhou2025learnable} & ICCV'25
& \textbf{0.465} & \underline{0.318} & \underline{0.235} & 0.182 & \underline{0.336} \\

\cmidrule(lr){2-8}
& \textbf{DRRG (0.6B)} & -
& \underline{0.452} & \textbf{0.332} & \textbf{0.259} &
\textbf{0.210} & \textbf{0.343} \\

\bottomrule
\end{tabular}%
}
\vspace{-1mm}
\end{table*}

\begin{table}[h]
\centering
\caption{Performance comparison of DRRG and existing RRG methods on CXPMRG-Bench for NLG and clinical metrics.}
\label{tab:chex_plus}
\small
\setlength{\tabcolsep}{4pt}
\resizebox{\columnwidth}{!}{%
\renewcommand{\arraystretch}{1}
\begin{tabular}{lccccc}
\toprule
\textbf{Model} & \textbf{BLEU-4} & \textbf{ROUGE-L} & \textbf{Precision} & \textbf{Recall} & \textbf{F1} \\
\midrule
XProNet~\cite{wang2022cross}        & 0.100 & 0.265 & 0.314 & 0.247 & 0.259 \\
ORGan~\cite{hou2023organ}          & 0.086 & 0.261 & 0.288 & 0.287 & 0.277 \\
M2KT~\cite{yang2023radiology}           & 0.078 & 0.247 & 0.044 & 0.142 & 0.058 \\
TIMER~\cite{wu2023token}          & 0.083 & 0.254 & 0.345 & 0.238 & 0.234 \\
CvT2DistilGPT2~\cite{nicolson2023improving} & 0.067 & 0.238 & 0.285 & 0.252 & 0.246 \\
R2Gen~\cite{chen2020generating}          & 0.081 & 0.246 & 0.318 & 0.200 & 0.181 \\
R2GenCMN~\cite{chen2022cross}       & 0.087 & 0.256 & 0.329 & 0.241 & 0.231 \\
Zhu et al.~\cite{zhu2023utilizing}     & 0.074 & 0.235 & 0.217 & 0.308 & 0.205 \\
CAMANet~\cite{wang2024camanet}        & 0.083 & 0.249 & 0.328 & 0.224 & 0.216 \\
Token-Mixer~\cite{yang2024token}    & 0.091 & 0.261 & 0.309 & 0.270 & 0.288 \\
PromptMRG~\cite{jin2024promptmrg}      & 0.095 & 0.222 & 0.258 & 0.265 & 0.281 \\
R2GenGPT~\cite{wang2023r2gengpt}       & 0.101 & 0.266 & 0.315 & 0.244 & 0.260 \\
R2GenCSR~\cite{wang2024r2gencsr}       & 0.100 & 0.265 & 0.315 & 0.247 & 0.259 \\
MambaXray-VL-B~\cite{wang2025cxpmrg} & 0.105 & \underline{0.267} & 0.333 & 0.264 & 0.273 \\
MambaXray-VL-L~\cite{wang2025cxpmrg} & \underline{0.112} & \textbf{0.276} & \textbf{0.377} & \underline{0.319} & \underline{0.335} \\
\textbf{DRRG}           & \textbf{0.119} & 0.265 & \underline{0.374} & \textbf{0.342} & \textbf{0.347} \\
\bottomrule
\end{tabular}
}
\vspace{-1mm}
\end{table}

\begin{table}[!b]
\centering
\caption{Results for CheXpert Precision, Recall, and F1 scores on the MIMIC-CXR dataset.}
\label{tab:performance_CheXbert}
\scriptsize
\setlength{\tabcolsep}{3pt}
\renewcommand{\arraystretch}{1}
\resizebox{\columnwidth}{!}{%
\begin{tabular}{llccc}
\toprule
\textbf{Model} & \textbf{Publication} & \textbf{Precision} & \textbf{Recall} & \textbf{F1} \\
\midrule
R2Gen~\cite{chen2020generating} & EMNLP'20 & 0.333 & 0.273 & 0.276 \\
R2GenCMN~\cite{chen2022cross} & ACL-IJCNLP'21 & 0.334 & 0.275 & 0.278 \\
R2GenGPT~\cite{wang2023r2gengpt} & Meta-Rad'23 & 0.392 & 0.387 & 0.389 \\
METransformer~\cite{wang2023metransformer} & CVPR'23 & 0.364 & 0.309 & 0.311 \\
Multi-Grained~\cite{liu2024multi} & TMI'24 & 0.457 & 0.337 & 0.330 \\
MambaXray-VL-Large~\cite{wang2025cxpmrg} & CVPR'25 & 0.411 & 0.373 & 0.395 \\
KACL~\cite{sha2025contrastive} & MICCAI'25 & 0.503 & 0.442 & 0.469 \\
DAMPER~\cite{huang2025damper} & AAAI'25 & 0.512 & 0.473 & 0.507 \\
DART~\cite{Park_2025_CVPR} & CVPR'25 & \underline{0.533} & \textbf{0.546} & \underline{0.520} \\
\textbf{DRRG} & - & \textbf{0.558} & \underline{0.540} & \textbf{0.549} \\
\bottomrule
\end{tabular}%
}
\end{table}

\subsubsection{Inference Process}
At inference time, we follow the iterative denoising procedure of LLaDA~\cite{nie2026large}. Given a target length $L$, we initialize the report as a fully masked sequence $G_1$ and iteratively denoise and refine it over $K$ discrete timesteps until obtaining the clean output prediction $\hat{G}_0$.

At each timestep, the mask ratio is $t_i$; the denoising model simultaneously predicts token distributions for all masked positions, conditioned on the concept-conditioned features $H$, clinical context $Y$, and current sequence $G_{t_i}$. For each masked position $j$, we select the most likely token
\begin{equation}
\hat{g}[j]=\arg\max_{v\in\mathcal{V}}\,p_{\theta}(v\mid H, Y, G_{t_i}),
\end{equation}
and compute the confidence score. $\mathcal{V}$ denotes the vocabulary. Following the low-confidence re-masking strategy, we rank all masked positions by confidence and commit only the top $m_i$ newly committed tokens specified by the $K$-step schedule. The remaining low-confidence positions are re-masked and refined in the next timestep. Let $\Omega_i$ denote the selected high-confidence positions. The update is:
\begin{equation}
G_{t_{i+1}}[j] \;=\;
\begin{cases}
\hat{g}[j], & j \in \Omega_i,\\
\texttt{[MASK]}, & j \in \mathcal{M}_{t_i} \setminus \Omega_i,
\end{cases}
\qquad |\Omega_i| = m_i .
\end{equation}

\section{Results}
Our model adopts a vision--language architecture with a SigLIP2 vision encoder~\cite{tschannen2025siglip} and a Qwen3-0.6B-diffusion-mdlm-v0.1 text decoder~\cite{zhou2026dllm}. We fully fine-tune the model on two NVIDIA RTX 5090 GPUs using a per-device batch size of 2, gradient accumulation over 4 steps, and an initial learning rate of $1\times10^{-4}$. AdamW is used with a cosine learning rate schedule and a warmup ratio of 0.03. During evaluation, we set the sequence length to $L=100$ and set the number of denoising steps to $K=72$.

To comprehensively evaluate our model, we adopt three categories of metrics: lexical, clinical, and LLM-based metrics. For lexical evaluation, we employ widely used natural language generation (NLG) metrics, including BLEU~\cite{papineni2002bleu} and ROUGE-L~\cite{lin2004rouge}, to measure surface-level similarity between generated and reference reports. For clinical evaluation, we report CheXpert precision, recall, and F1 scores~\cite{smit2020chexbert}, which assess report-level agreement across 14 CheXpert observations~\cite{irvin2019chexpert}. We further use RadGraph-F1~\cite{jain2021radgraph} to evaluate consistency at the clinical entity and relation levels. To provide an assessment more closely aligned with expert clinical judgement, we additionally adopt two LLM-based metrics. GREEN~\cite{ostmeier2024green} identifies clinically meaningful errors, whereas RaTEScore~\cite{zhao2024ratescore} measures clinical entity correspondence between generated and reference reports while accounting for synonyms and negation.

Tables~\ref{tab:mimic_nlg}, \ref{tab:chex_plus}, \ref{tab:performance_CheXbert}, and~\ref{tab:performance_llm} compare DRRG with existing RRG methods. Specifically, Table~\ref{tab:mimic_nlg} reports NLG performance on MIMIC-CXR, while Table~\ref{tab:chex_plus} presents both NLG and CheXpert-based clinical results on CXPMRG-Bench. Tables~\ref{tab:performance_CheXbert} and~\ref{tab:performance_llm} further evaluate clinical reliability on MIMIC-CXR using CheXpert precision, recall, and F1, as well as RadGraph-F1, GREEN, and RaTEScore.

\begin{figure}[!t]
    \centering
    \includegraphics[width=1\linewidth]{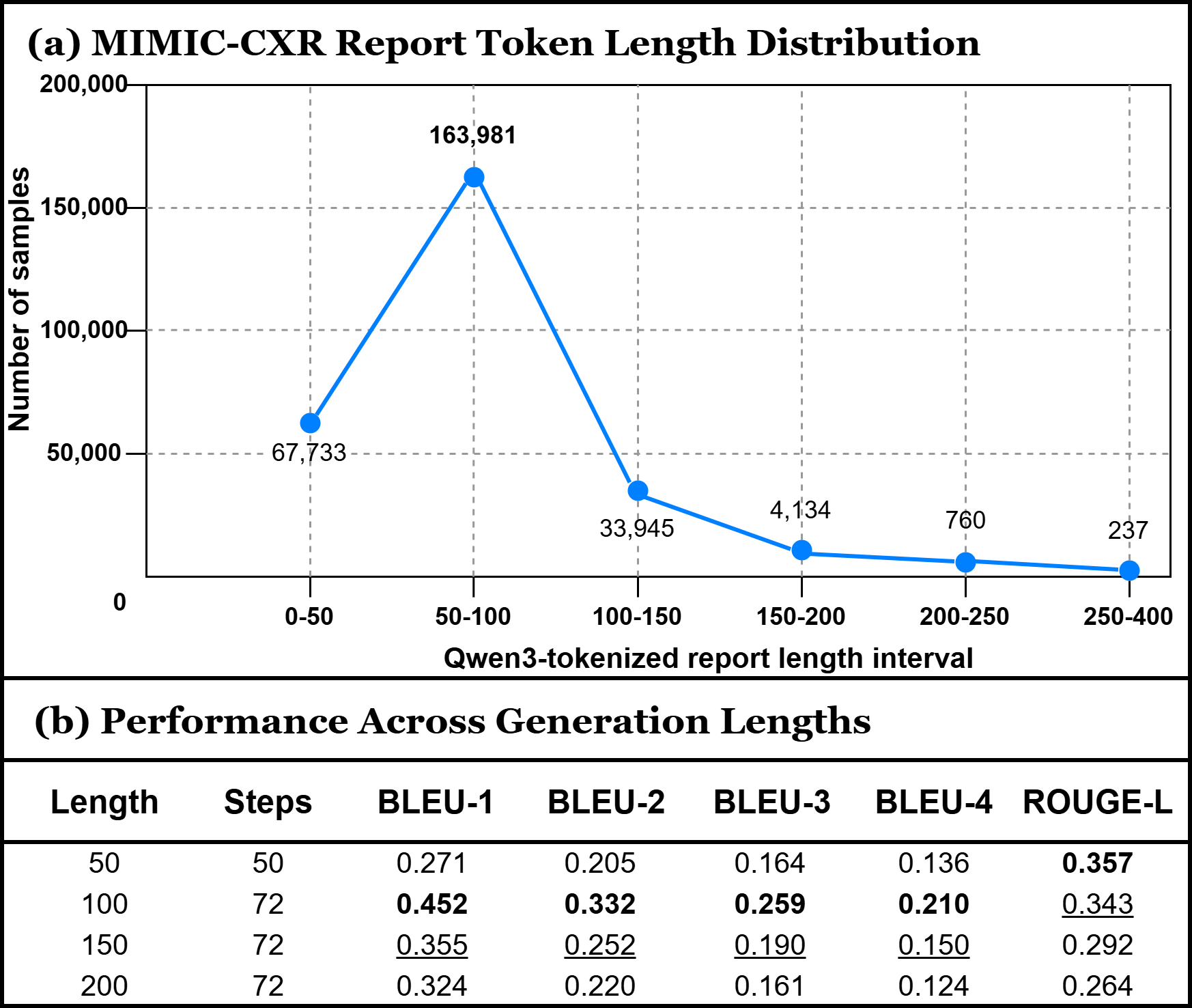}
    \caption{(a) Distribution of Qwen3-tokenized report lengths in MIMIC-CXR. (b) Performance under different initial generation lengths.}
    \label{fig:dataset}
\end{figure}

\begin{figure}[!t]
    \centering
    \includegraphics[width=\linewidth]{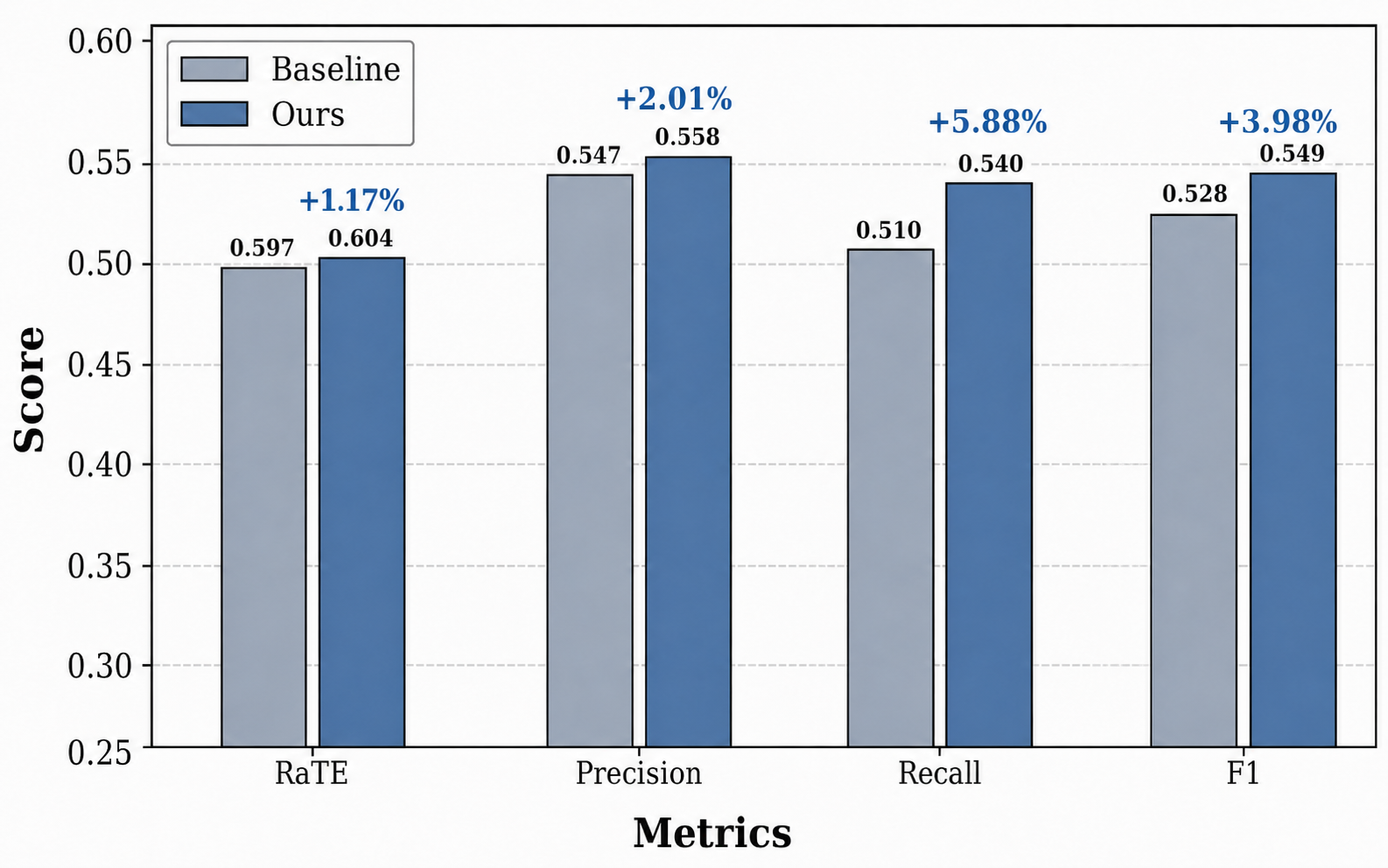}
    \caption{Clinical metric and LLM-based metric results of the ablation study, comparing the performance of the baseline and DRRG models.}
    \label{fig:abalation}
\end{figure}

Tables~\ref{tab:ablation_components} and \ref{tab:diff_steps} and Figures~\ref{fig:dataset}, \ref{fig:factor}, \ref{fig:a_factor}, and~\ref{fig:abalation} present the ablation and hyperparameter analyses. Table~\ref{tab:ablation_components} evaluates the contributions of the diffusion framework, clinical-entities-aware complementary mask, and concept-conditioning module, while Figure~\ref{fig:abalation} reports their effects on clinical and LLM-based metrics. Figures~\ref{fig:a_factor}, \ref{fig:factor}, and~\ref{fig:dataset} examine the effects of the clinical-entity masking factor, concept-loss weight, and initial generation length, respectively. Table~\ref{tab:diff_steps} further compares the performance and inference efficiency of autoregressive and diffusion decoding under different denoising steps.

\begin{figure*}[t]
    \centering
    \includegraphics[width=\linewidth]{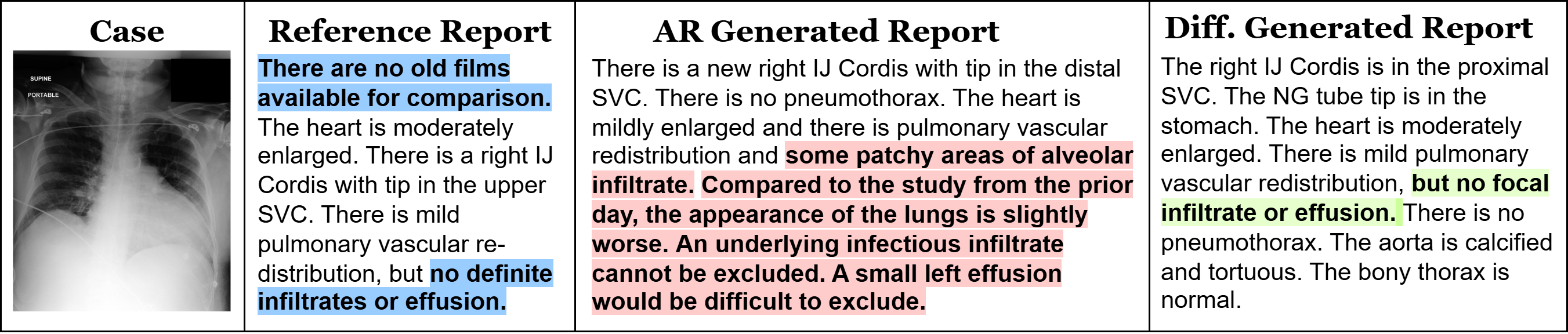}
    \caption{Case study comparing reports generated by AR and DLLM models. Words highlighted in red indicate incorrect descriptions, words highlighted in blue denote reference content, and words highlighted in green indicate correct descriptions.}
    \label{fig:case}
\end{figure*}

Figures~\ref{fig:case} and~\ref{fig:case2} provide qualitative case studies. Figure~\ref{fig:case} compares reports generated by the AR and DLLM models to illustrate their different error patterns, whereas Figure~\ref{fig:case2} visualizes the iterative refinement process and shows how additional denoising steps correct incorrect descriptions.

\begin{table}[!t]
\centering
\caption{Results on the LLM-based metrics GREEN and RaTEScore, as well as the clinical metric RadGraph-F1 (RG$_{F1}$).}
\label{tab:performance_llm}
\scriptsize
\setlength{\tabcolsep}{3pt}
\renewcommand{\arraystretch}{1}
\resizebox{\columnwidth}{!}{%
\begin{tabular}{llccc}
\toprule
\textbf{Model} & \textbf{Publication} & \textbf{RG$_{F1}$} & \textbf{GREEN} & \textbf{RaTE} \\
\midrule
R2Gen~\cite{chen2020generating} & EMNLP'20 & 0.172 & 0.276 & 0.526 \\
R2GenCMN~\cite{chen2022cross} & ACL-IJCNLP'21 & 0.182 & 0.297 & \underline{0.538} \\
CvT2DistilGPT2~\cite{nicolson2023improving} & AIM'23 & 0.196 & \underline{0.320} & 0.527 \\
R2GenGPT~\cite{wang2023r2gengpt} & Meta-Rad'23 & 0.187 & 0.300 & 0.528 \\
PromptMRG~\cite{jin2024promptmrg} & AAAI'24 & 0.190 & 0.287 & 0.528 \\
KARGEN~\cite{li2024kargen} & ICMCC'24 & 0.203 & 0.308 & 0.533 \\
EKAGen~\cite{bu2024instance} & CVPR'24 & 0.199 & 0.256 & 0.512 \\
MultiP-R2Gen~\cite{chen2025enhancing} & TMI'25 & \underline{0.208} & -- & -- \\
\textbf{DRRG} & - & \textbf{0.281} & \textbf{0.360} & \textbf{0.604} \\
\bottomrule
\end{tabular}%
}
\end{table}

\section{Discussion}

\subsection{Comparison with Existing Methods}
To better assess the performance of DRRG, we compare our model with existing RRG frameworks on the MIMIC-CXR and CheXpert Plus datasets. We report the results using NLG, clinical, and LLM-based metrics to provide a comprehensive evaluation of the model.

\textbf{NLG Metrics:} The NLG results on MIMIC-CXR and CheXpert Plus are reported in Table~\ref{tab:mimic_nlg} and Table~\ref{tab:chex_plus}. On MIMIC-CXR, DRRG achieves the best performance on BLEU-2, BLEU-3, BLEU-4, and ROUGE-L, with scores of 0.332, 0.259, 0.210, and 0.343, respectively. It also obtains the second-best BLEU-1 score of 0.452, slightly behind REVTAF-RRG~\cite{zhou2025learnable}. Compared with recent LLM-based RRG methods, DRRG achieves stronger NLG performance with a substantially smaller 0.6B text decoder, whereas the listed LLM-based competitors use text decoders with 7B or more parameters. On CheXpert Plus, DRRG achieves the best BLEU-4 score and a competitive ROUGE-L of 0.265. These results demonstrate the effectiveness of DRRG, suggesting that DLLMs provide a strong alternative to conventional AR-based RRG models.

\begin{table}[!htbp]
\centering
\caption{Ablation study conducted on the MIMIC-CXR dataset. “Diff.” indicates whether a DLLM or an AR model is used. “Clinical-entities” indicates whether clinical entity tokens are assigned higher masking probabilities than other report tokens.
“Concept-condition” indicates whether concept features are added to the vision features.}
\label{tab:ablation_components}
\scriptsize
\setlength{\tabcolsep}{2pt}
\renewcommand{\arraystretch}{1.05}
\resizebox{\columnwidth}{!}{%
\begin{tabular}{ccc|ccccc}
\toprule
\textbf{Diff.} & \textbf{Clinical-entities} & \textbf{Concept-condition} 
& \textbf{BLEU-1} & \textbf{BLEU-2} & \textbf{BLEU-3} & \textbf{BLEU-4} & \textbf{ROUGE-L} \\
\midrule
$\times$ & $\times$ & $\times$ & 0.430 & 0.306 & 0.236 & 0.192 & 0.300 \\
$\checkmark$ & $\times$ & $\times$ & 0.449 & 0.328 & 0.254 & 0.204 & 0.335 \\
$\checkmark$ & $\checkmark$ & $\times$ & \underline{0.450} & \underline{0.330} & \underline{0.256} & \underline{0.206} & \underline{0.340} \\
$\checkmark$ & $\times$ & $\checkmark$ & 0.445 & 0.326 & 0.254 & 0.205 & 0.338 \\
$\checkmark$ & $\checkmark$ & $\checkmark$ & \textbf{0.452} & \textbf{0.332} & \textbf{0.259} & \textbf{0.210} & \textbf{0.343} \\
\bottomrule
\end{tabular}%
}
\end{table}

\begin{table}[!h]
\centering
\caption{Performance and efficiency comparison of autoregressive decoding and diffusion decoding with different denoising steps.}
\label{tab:diff_steps}
\scriptsize
\setlength{\tabcolsep}{2pt}
\renewcommand{\arraystretch}{1.05}
\resizebox{\columnwidth}{!}{%
\begin{tabular}{cccccccc}
\toprule
\textbf{Diff.} & \textbf{Steps} & \textbf{BLEU-1} & \textbf{BLEU-2} & \textbf{BLEU-3} & \textbf{BLEU-4} & \textbf{ROUGE-L} & \textbf{Time (case)} \\
\midrule
$\times$ & -- & 0.430 & 0.306 & 0.236 & 0.192 & 0.300 & 1.80s \\
$\checkmark$ & 32 & 0.447 & 0.325 & 0.250 & 0.200 & 0.338 & 0.97s \\
$\checkmark$ & 64 & \underline{0.451} & \underline{0.331} & \underline{0.258} & \underline{0.208} & \underline{0.342} & 1.91s \\
$\checkmark$ & 72 & \textbf{0.452} & \textbf{0.332} & \textbf{0.259} & \textbf{0.210} & \textbf{0.343} & 2.13s \\
\bottomrule
\end{tabular}
}
\end{table}

\textbf{Clinical Metrics:} Table~\ref{tab:chex_plus}, Table~\ref{tab:performance_CheXbert}, and Table~\ref{tab:performance_llm} report the clinical performance of DRRG and competing methods on CheXpert Plus and MIMIC-CXR. On CheXpert Plus, DRRG achieves the best CheXpert recall and F1 scores, outperforming MambaXray-VL-L~\cite{wang2025cxpmrg} by 0.023 and 0.012, respectively, while obtaining the second-best precision score of 0.374. On MIMIC-CXR, DRRG achieves the best precision and F1 scores of 0.558 and 0.549, surpassing DART~\cite{Park_2025_CVPR} by 0.025 and 0.029, respectively, with only a marginally lower recall. In addition, DRRG obtains a RadGraph-F1 score of 0.281, which is 35.1\% higher than the second-best MultiP-R2Gen~\cite{chen2025enhancing}. These results show that DRRG not only improves textual similarity but also better preserves clinically relevant findings, demonstrating its effectiveness for radiology report generation.

\textbf{LLM-based Metrics:} As shown in Table~\ref{tab:performance_llm}, DRRG further achieves the best LLM-based evaluation results, with GREEN and RaTEScore values of 0.360 and 0.604, respectively. This suggests that its generated reports are more consistent with clinically grounded assessment criteria.

 \begin{figure*}[!t]
    \centering
    \includegraphics[width=\linewidth]{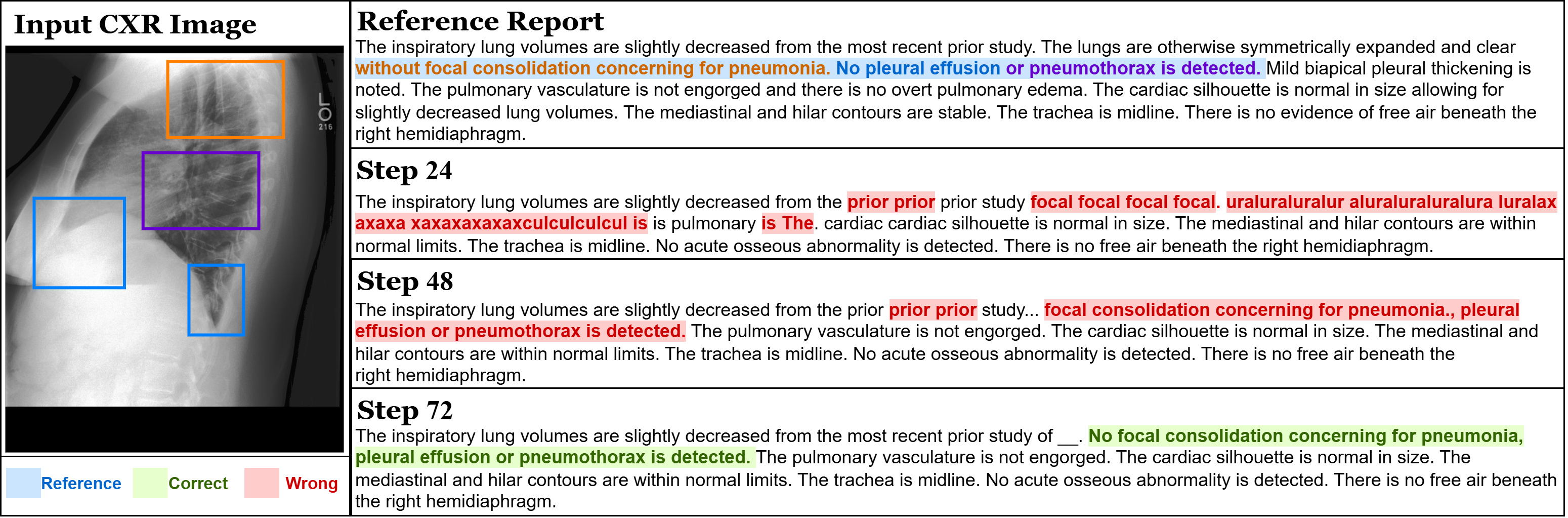}
    \caption{Case study of DRRG’s iterative refinement. Reference colors indicate the corresponding image regions, while blue, green, and red highlights denote reference, correct, and incorrect findings, respectively. Step 72 further refines the report by correcting false findings generated at Step 48.
}
    \label{fig:case2}
\end{figure*}

 \begin{figure}[!t]
    \centering
    \includegraphics[width=\linewidth]{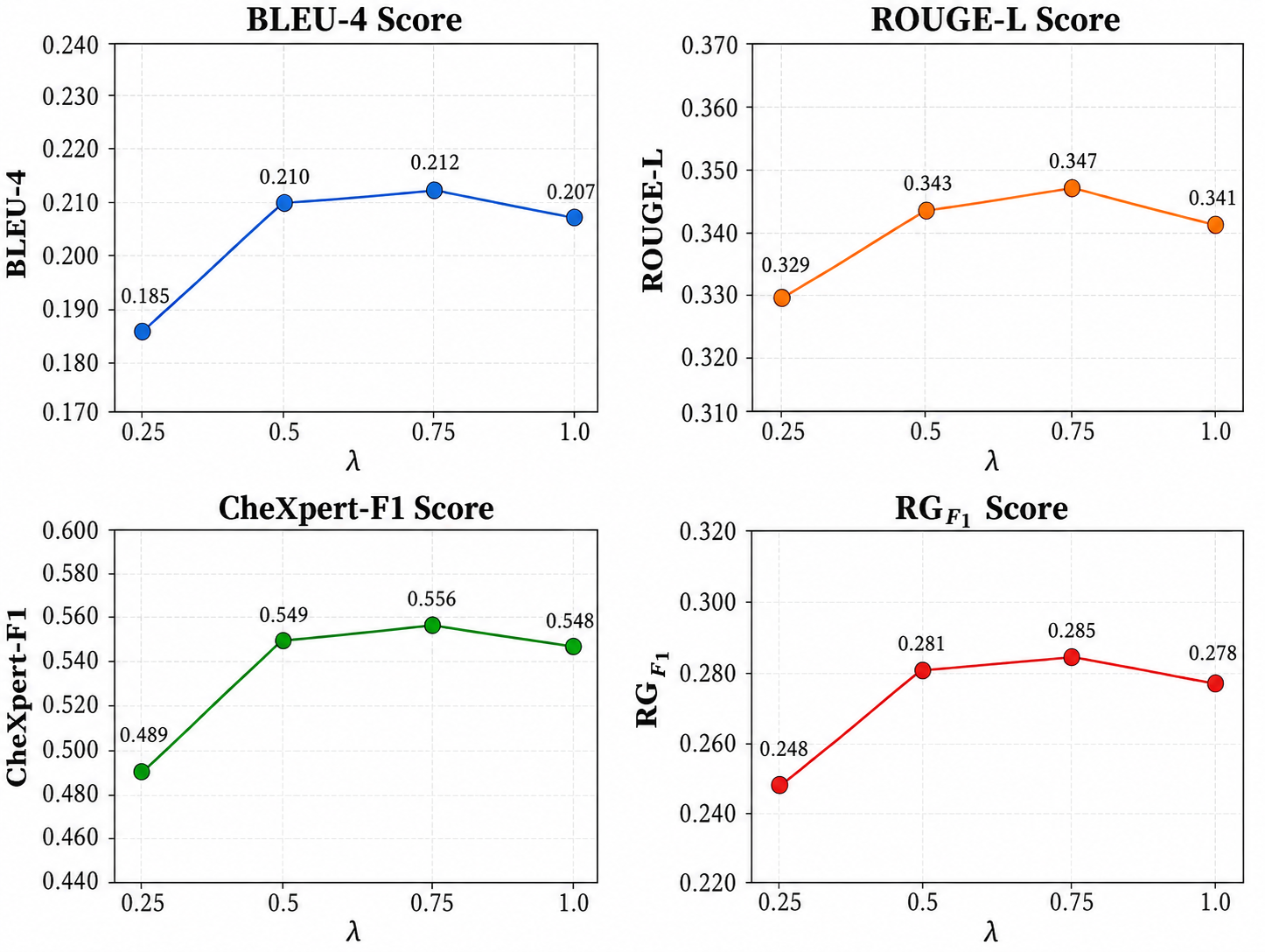}
    \caption{Effect of the weighting factor $\lambda$ for the concept loss $\mathcal{L}_c$ on model performance in terms of BLEU-4, ROUGE-L, CheXpert-F1, and RadGraph-F1 (RG$_{F1}$).}
    \label{fig:factor}
\end{figure}

 \begin{figure}[!t]
    \centering
    \includegraphics[width=\linewidth]{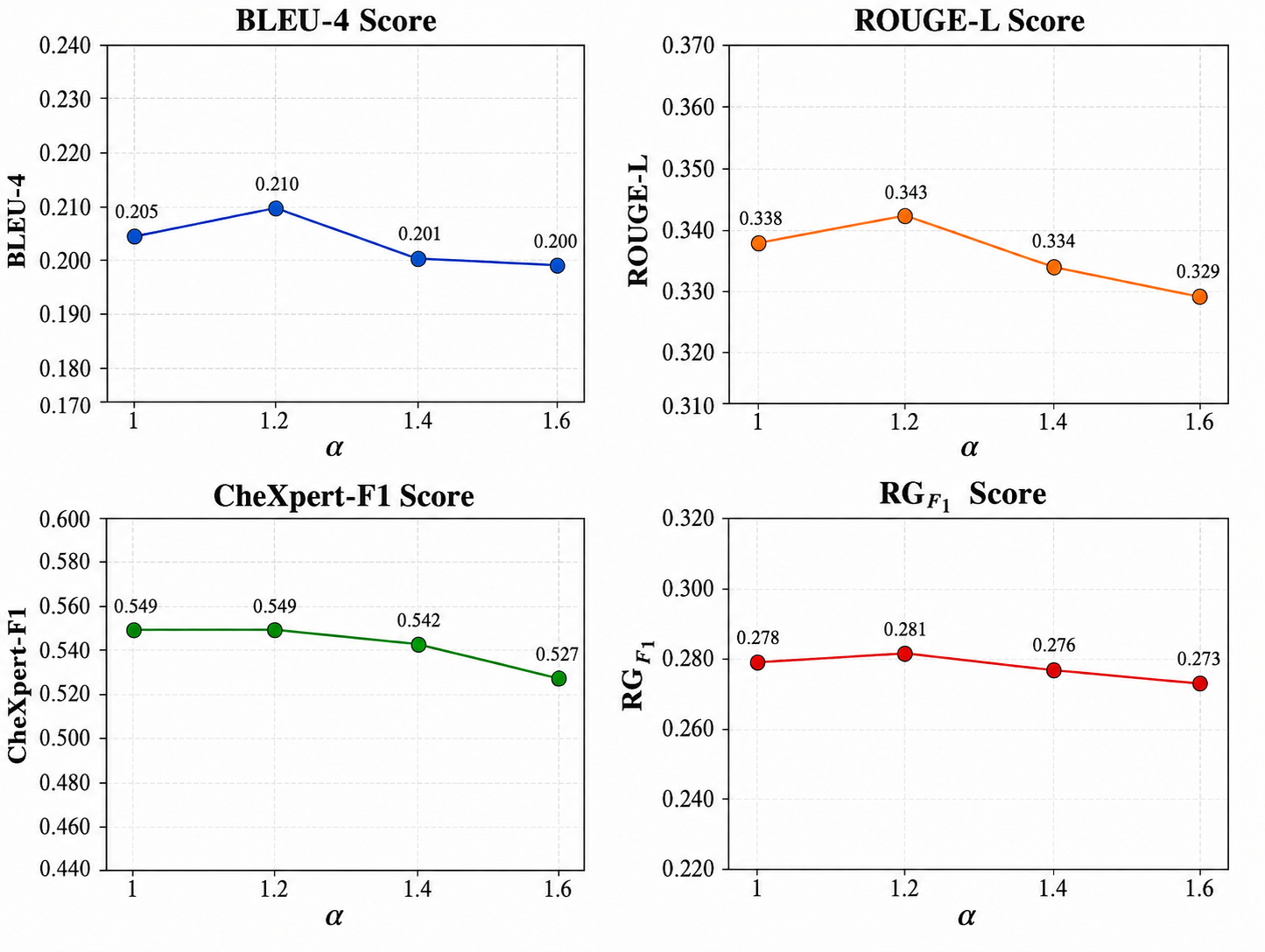}
    \caption{Effect of the clinical-entity masking factor probability factor $\alpha$ in terms of BLEU-4, ROUGE-L, CheXpert-F1, and RadGraph-F1 (RG$_{F1}$).}
    \label{fig:a_factor}
\end{figure}

\subsection{Ablation Analysis of Model Components}
We ablate the key components of DRRG on MIMIC-CXR using different metrics. The results are reported in Table~\ref{tab:ablation_components} and Figure~\ref{fig:abalation}. We further report the performance--efficiency comparison between DRRG and the AR baseline across different denoising steps in Table~\ref{tab:diff_steps}.

We first compare AR- and DLLM-based generation paradigms under comparable model configurations. Specifically, the AR baseline uses the same SigLIP2 vision encoder~\cite{tschannen2025siglip} and a Qwen3-0.6B text decoder~\cite{yang2025qwen3}, making the training paradigm the main difference between the two models. The DLLM variant is equipped with the standard complementary mask~\cite{li2025lavida}, enabling all tokens in each sample to contribute to training. As shown in Table~\ref{tab:ablation_components}, the DLLM-based model consistently outperforms the AR baseline across all NLG metrics, with gains of 0.019, 0.022, 0.018, 0.012, and 0.035, respectively. These results demonstrate the effectiveness of DLLM-based iterative denoising for radiology report generation under the evaluated setting.

We further analyze the contribution of each DRRG module. Introducing clinical-entities-aware probability adjustment into the standard complementary masking strategy yields an average NLG improvement of 0.0024 across the five NLG metrics, suggesting its benefit for fine-grained clinical entity modeling. When the concept-conditioning module is further incorporated, the two modules bring larger gains across both NLG and clinical metrics, improving BLEU-1, BLEU-2, BLEU-3, BLEU-4, and ROUGE-L by 0.003, 0.004, 0.005, 0.006, and 0.008, respectively. RaTEScore, CheXpert precision, recall, and F1 further increase by 1.17\%, 2.01\%, 5.88\%, and 3.98\%. These improvements indicate that both designs enhance report quality and clinical consistency.

\subsection{Analysis of Training and Generation Settings}
Beyond evaluating the contribution of individual model components, we further investigate the effects of key hyperparameters in both the training and generation stages, including the masking ratio of clinical entities, the weighting factor of the concept loss, the initial generation length, and the number of denoising steps.

To examine the effect of the clinical-entity masking factor, we vary $\alpha$ from $1.0$ to $1.6$, as reported in Figure~\ref{fig:a_factor}. Increasing $\alpha$ from $1.0$ to $1.2$ improves BLEU-4 from $0.205$ to $0.210$, ROUGE-L from $0.338$ to $0.343$, and RadGraph F1 from $0.278$ to $0.281$. These results indicate that moderately increasing the masking probability of core findings encourages the model to place greater emphasis on clinical entities. However, further increasing $\alpha$ leads to a decline across all evaluation metrics. A larger masking factor causes more clinical entities to be removed from the corrupted reports during training, thereby reducing the number of training instances in which the model can learn to recover masked content from the remaining visible clinical entities. Consequently, during iterative inference, the model becomes less capable of using already recovered clinical entities to reconstruct the remaining masked tokens. Overall, $\alpha=1.2$ provides the most appropriate balance between enhancing the model's attention to clinical entities and preserving sufficient entity-level information for iterative reconstruction.

Regarding the concept-loss weight, as shown in Figure~\ref{fig:factor}, assigning a small weight of 0.25 results in the lowest performance across all four evaluation metrics. Increasing the weight from 0.25 to 0.5 improves BLEU-4, ROUGE-L, CheXpert-F1, and RadGraph-F1 by 0.025, 0.014, 0.060, and 0.033, respectively, demonstrating that explicit supervision of clinical observations can effectively guide the denoising process and improve both linguistic quality and clinical correctness. The performance continues to improve slightly when the weight is increased to 0.75. However, further increasing it to 1.0 leads to consistent degradation across all metrics. This suggests that an excessively large concept-loss weight may overemphasize observation classification at the expense of coherent and accurate report generation.

As shown in Figure~\ref{fig:dataset}, the initial generation length has a substantial impact on model performance. An overly short sequence yields relatively poor performance on most metrics, as the restricted generation capacity may force the model to omit clinically important findings. The best overall performance is achieved at an initial length of 100 tokens, which provides sufficient capacity for comprehensive report generation while closely matching the empirical training distribution: 231,714 of the 270,790 training samples contain fewer than 100 tokens. In contrast, sequences of 150 and 200 tokens are sparsely represented during training, limiting the model's ability to learn reliable denoising dynamics for such long sequences. In addition, with the number of denoising steps fixed, a longer initial sequence requires more tokens to be retained at each iteration. Consequently, some low-confidence predictions may be preserved prematurely instead of being further refined, increasing the likelihood that erroneous content remains in the final report. These two factors jointly account for the performance degradation observed at longer initial generation lengths. Nevertheless, enabling discrete diffusion models to adaptively determine the appropriate generation length remains an open problem that is being actively explored in recent studies~\cite{li2026beyond,liu2026diffusion,yang2025diffusion}.

For the denoising steps, increasing the steps from 32 to 72 consistently improves all NLG metrics, indicating that more denoising iterations enable better report refinement, but also increase inference time from 0.97s to 2.13s per case. Compared with the autoregressive baseline, DRRG achieves a better quality--efficiency trade-off: with 64 steps, it obtains higher scores at a comparable runtime, and even with only 32 steps, it still outperforms the corresponding AR baseline while using only 53.9\% of its inference time. These results demonstrate the effectiveness and efficiency of DRRG.

\subsection{Qualitative Analysis}
To examine the potential of DLLMs in mitigating error propagation, we present a case study in Figure~\ref{fig:case}, which compares the reports generated by the AR-based model and the DLLM-based model with the reference report. The AR-generated report hallucinates several findings, beginning with the early erroneous phrase “some patchy areas of alveolar infiltrate.” This illustrates a typical error-propagation issue in AR models, where an early incorrect token sequence is fixed as context and further amplifies subsequent hallucinations. In contrast, the DLLM-generated report better matches the reference report. Through iterative denoising, low-confidence tokens can be re-masked and revised, thereby reducing the impact of error propagation and improving report accuracy.

Figure~\ref{fig:case2} illustrates the progressive refinement process of DRRG. The reference report describes several key findings as absent, including focal consolidation, pleural effusion, and pneumothorax. At Step 24, the generated report remains dominated by noisy intermediate predictions, with numerous duplicated and fragmented tokens, while clinically meaningful findings have not yet been reliably recovered. By Step 48, the sentence structure has become substantially more coherent; however, the model fails to generate the appropriate negation and incorrectly converts several absent findings into positive observations, resulting in a potentially clinically consequential false-positive statement. With additional denoising steps, the report at Step 72 restores the correct negation for consolidation, pleural effusion, and pneumothorax, while further improving fluency by reducing grammatical artifacts. This example illustrates the potential of iterative denoising to correct clinically important semantic errors and progressively improve report quality.

\section{Conclusion}
We proposed DRRG, a discrete diffusion framework for radiology report generation. By formulating report generation as iterative denoising rather than autoregressive decoding, DRRG enables bidirectional refinement and better aligns with the revisable nature of radiology reporting. With the clinical-entities-aware complementary mask and concept-conditioning module, DRRG improves data efficiency, enhances clinical-entity awareness, and guides the denoising process with clinical concepts. Extensive experiments on MIMIC-CXR and CheXpert Plus demonstrate its effectiveness across different evaluation metrics. Further ablation studies validate the contribution of the proposed modules, while the case study illustrates the potential of DRRG to mitigate error propagation in report generation.

\printcredits

\section*{Ethics statement}
This study used two publicly available and de-identified chest radiography datasets, MIMIC-CXR and CheXpert Plus. No new participants were recruited, and no identifiable personal information was collected or accessed in this study. All data were used in accordance with the corresponding data-use requirements and institutional access procedures. Therefore, this study constituted a secondary analysis of existing de-identified data and did not require additional informed consent from individual patients.

\section*{Availability of data and materials}
The datasets used in this study, including MIMIC-CXR and CheXpert Plus, are publicly available from their respective official repositories, subject to the corresponding access requirements and data-use agreements. The source code and implementation details of DRRG will be made publicly available on GitHub upon publication.

\section*{Declaration of generative AI and AI-assisted technologies in the writing process}
During the preparation of this work, the authors used ChatGPT (OpenAI) to improve the readability and language of the manuscript. After using this tool, the authors carefully reviewed and edited the content as needed and take full responsibility for the content of the published article.

\section*{Declaration of competing interests}
Luping Zhou is a member of the Editorial Board of Meta-Radiology. She was not involved in the peer-review process. The manuscript was independently handled by another member of the Editorial Board. All authors declare that they have no competing financial interests or personal relationships that could have influenced the work reported in this paper.

\section*{Acknowledgements}
The authors have no acknowledgements to declare.
\bibliographystyle{cas-model2-names}
\bibliography{cas-refs}


\end{document}